\documentclass{article} 
\usepackage{iclr2021_conference,times}

\usepackage{amsmath,amsfonts,bm}

\def\eqref#1{equation~\ref{#1}}

\def\1{\bm{1}}

\DeclareMathAlphabet{\mathsfit}{\encodingdefault}{\sfdefault}{m}{sl}
\SetMathAlphabet{\mathsfit}{bold}{\encodingdefault}{\sfdefault}{bx}{n}

\usepackage{hyperref}
\usepackage{url}

\usepackage{bbm}
\usepackage{multirow}
\usepackage{amsmath,amssymb,amsfonts}
\usepackage{graphicx}
\usepackage{booktabs}
\usepackage{wrapfig}
\usepackage{hhline}
\usepackage{appendix}
\usepackage{caption}
\usepackage{subcaption}
\usepackage{csquotes}
\usepackage[greek,british]{babel}
\usepackage{enumitem}
\usepackage[bottom]{footmisc}

\usepackage{adjustbox}
\usepackage{color, colortbl}
\usepackage{tikz,lipsum,lmodern}
\usepackage[most]{tcolorbox}
\definecolor{tableblue}{rgb}{0.871,0.922,0.969}

\usepackage[utf8x]{inputenc}

\usepackage{caption}
\usepackage{capt-of}
\usepackage{subcaption}

\title{Let Your Heart Speak in its Mother Tongue: Multilingual Captioning of Cardiac Signals}

\author{Dani Kiyasseh, Tingting Zhu \& David Clifton \\
Department of Engineering Science\\
University of Oxford\\
Oxford, UK \\
\texttt{\{dani.kiyasseh,tingting.zhu,david.clifton\}@eng.ox.ac.uk} \\
}

\iclrfinalcopy 
\begin{document}

\maketitle

\begin{abstract}
Cardiac signals, such as the electrocardiogram, convey a significant amount of information about the health status of a patient which is typically summarized by a clinician in the form of a clinical report, a cumbersome process that is prone to errors. To streamline this routine process, we propose a deep neural network capable of captioning cardiac signals; it receives a cardiac signal as input and generates a clinical report as output. We extend this further to generate multilingual reports. To that end, we create and make publicly available a multilingual clinical report dataset. In the absence of sufficient labelled data, deep neural networks can benefit from a \enquote{warm-start}, or pre-training, procedure in which parameters are first learned in an arbitrary task. We propose such a task in the form of discriminative multilingual pre-training where tokens from clinical reports are randomly replaced with those from other languages and the network is tasked with predicting the language of all tokens. We show that our method performs on par with state-of-the-art pre-training methods such as MLM, ELECTRA, and MARGE, while simultaneously generating diverse and plausible clinical reports. We also demonstrate that multilingual models can outperform their monolingual counterparts, informally terming this beneficial phenomenon as the \enquote{blessing of multilinguality}.
\end{abstract}

\section{Introduction}

Cardiac signals, such as the electrocardiogram (ECG), convey a significant amount of information about a patient's clinical state. Upon recording these signals, an expert cardiologist is often required to interpret the findings, generate an accompanying textual report, and share it with fellow physicians (or patients) to complement the rest of a patient's medical results during a hospital visit. For an ECG to be of maximal value, clinical guidelines stipulate that it must be accompanied by a textual report \citep{SCST2020}. Such reports, however, can be time-consuming to generate and thus detract physicians from caring for patients. They also exhibit a high degree of ambiguity and inter/intra-physician variability which erodes patient-physician communication \citep{Hibbard2001,Keselman2012}. Automating the generation of ECG reports can streamline the clinical workflow for cardiologists, allow for more consistent ECG interpretations \citep{Willems1991}, and potentially reduce the 'inadvertent oversight' of medical conditions \citep{Brailer1997}

Notable progress has been achieved at the intersection of computer vision and natural language processing (NLP) on tasks such as image captioning (IC) \citep{Herdade2019,Lu2019}, visual question answering (VQA) \citep{Goyal2017,Anderson2018}, and text-to-video retrieval \citep{Miech2019}. In IC, for example, the goal is to generate text that describes the content of an image or the activity taking place in a video. This goal has been established within the medical imaging community \citep{Hasan2018,Kisilev2016,Zeng2020,Liu2019,Wang2018,Yoon2019}. However, the research community has not explored captioning in the context of cardiac signals, and particularly not with ECG signals. This could be due to the lack of publicly-available datasets comprising cardiac signals and paired textual reports. The limited amount of such data impedes the ability of NLP systems to learn language representations that generalize well. This obstacle is commonly addressed via language pre-training tasks that are either generative, such as masked language modelling \citep{Lu2020,Singh2020}, or discriminative \citep{Clark2020}. Recent work has also illustrated the benefit of multilingual pre-training relative to its monolingual counterpart in solving NLP tasks \citep{Conneau2019,Pratap2020,Conneau2020,Huang2019Unicoder,Artetxe2020}. 

\textbf{Contributions.} Motivated by the aforementioned findings, our contributions are as follows:
\begin{enumerate}[leftmargin=0.5cm]
    \item \textbf{Multilingual ECG reports.} We translate reports paired with physiological signals into seven different languages, making it the first of its kind, and open-source them to facilitate research into multilingual captioning of cardiac signals.
    \item \textbf{Multilingual captioning of cardiac signals.} We design a captioning system that receives, as input, a cardiac signal and generates, as output, a clinical textual report, in multiple languages.
    \item \textbf{Replaced token language prediction.} We propose a multilingual discriminative language representation learning method that randomly selects tokens in a sequence, replaces them with tokens from different languages, and tasks a network with classifying the language of the tokens. 
\end{enumerate}

\section{Related Work}

\textbf{Visual and language representation learning.} Representation learning is an integral component of textual and visual systems. In the former, generative language representation learning tasks such as masked language modelling (MLM) havgproven effective \citep{Devlin2018,Liu2019Roberta}. These generative methods havgalso been extended to the multilingual case \citep{Conneau2019NIPS,Conneau2019,Liu2020}. Most similar to our work is ELECTRA \citep{Clark2020}, wherein tokens are replaced with those from a generative model, and a network is tasked with discriminating between the original and the replaced tokens. This approach simultaneously learns an MLM and discriminative network, deeming it computationally expensive and dependent on large datasets. In contrast, our work is not dependent on an additional MLM network and explicitly deals with the multilingual setting. MARGE \citep{Lewis2020} retrieves documents, in potentially different languages, and attempts to reconstruct a target document. Instead of focusing on a single modality, others propose to learn textual and visual representations jointly \citep{Sun2019,Lu2019,Zhang2020}. To the best of our knowledge, we are the first to propose a discriminative multilingual language representation learning method in the context of cardiac signals.

\textbf{Multilingual representation learning.} Pre-training and fine-tuning networks on multiple languages has been shown to benefit NLP tasks \citep{Conneau2019,Pratap2020,Conneau2020,Artetxe2020}. For example, \cite{Conneau2019} and \cite{Artetxe2020} show that multilingual pre-training is more advantageous than its monolingual counterpart when solving downstream NLP tasks. In some cases, this has been attributed to commonalities across languages such as word order, characters, and semantic structure. These findings havgbeen partially fueled by monolingual datasets which havgbeen machine-translated to multiple languages, such as XNLI \citep{Conneau2018}. In this work, we translate ECG reports into multiple languages and leverage that for pre-training and fine-tuning purposes. Most similar to our multilingual pre-training setup is that of \citep{Huang2019Unicoder}. Our work differs in that it explores more languages, defines a different pre-training task, and leverages that for cardiac captioning. 

\textbf{Image captioning in healthcare.} In the domain of biomedical IC, most research has focused on chest X-rays. For example, \cite{Hasan2018} propose to incorporate clinical concept prediction to improve the captioning performance. \cite{Kisilev2016,Zeng2020} introduce a multi-task objective to simultaneously perform bounding box regression and captioning. \cite{Liu2019} condition their captioning system on the medical topic to be discussed and \cite{Wang2018} propose a multi-level attention model that attends to both the image and the text. More recently, methods such as ClinicalBert \citep{Huang2019}, BioBert \citep{Lee2020}, and BioELMo \citep{Jin2019} were shown to learn rich representations of clinical text. These representations are also beneficial to biomedical applications \citep{Yoon2019}. To the best of our knowledge, we are the first to explore multilingual captioning in the context of cardiac signals.


\section{Background}

\subsection{Image Captioning}
\label{sec:image_captioning}
In image captioning, the goal is to generate a sequence of output words that reflect the content of the input signal. To extract features from the input signal, an encoder, $f_{\theta}: u \in \mathbb{R}^{D} \rightarrow v \in \mathbb{R}^{L \times M}$ parameterized by $\theta$, maps a $D$-dimensional input, $u$, to a set of $L$ representations, $v$, each of which are $M$-dimensional. Each of these representations corresponds approximately to a different receptive field of the input. 

In dealing with captions, we convert words to tokens (e.g., by lower-casing and stemming them) and form a fixed vocabulary, $V$, consisting of $C$ tokens, such that $|V|=C$. An embedding matrix (lookup table), $E \in \mathbb{R}^{C \times M}$, maps each of these $C$ tokens to an $M$-dimensional embedding, $e \in \mathbb{R}^{M}$. We can now represent any given caption as a sequence of token embeddings, $e \in \mathbb{R}^{M \times S}$, of length, $S$. We feed these token embeddings to a decoder, $g_{\phi}: v, e \in \mathbb{R}^{M \times S} \rightarrow h \in \mathbb{R}^{M \times S}$ parameterized by $\phi$, which at each step in the sequence attends to all the encoder representations, $v$, (e.g., via cross-attention) and generates an $M$-dimensional representation, $h$. These representations, $h$, are fed to a multi-layer perceptron (MLP) head, $p_{\omega} : h \in \mathbb{R}^{M \times S} \rightarrow y \in \mathbb{R}^{C \times S}$, to output a probability distribution over the $C$ tokens in the vocabulary at each step in the sequence. By identifying the most probable token at each step, we can form a sentence of words. 

\section{Methods}

\subsection{Multilingual Captioning of Cardiac Signals}
In cardiac signal captioning, the goal is to generate a caption of diagnostic value that reflects the content of a cardiac signal representing the clinical state of the patient. Given cardiac signals, $u$, and their associated captions, $t$, in a a single language, $l$, we define a dataset as $\mathcal{D}_{l}=\{u,t\}$. It would make most sense for a captioning system to be monolingual and in the mother tongue of the corresponding healthcare institution in which the system is deployed. However, generating captions in a single language implies that a separate model is needed for each language, an approach which is cumbersome to train and deploy \citep{Conneau2020}. Moreover, as described earlier, incorporating multilinguality into the training procedure (both during pre-training and fine-tuning) has been shown to benefit downstream NLP tasks \citep{Artetxe2020}.

\begin{figure}[!h]
    \centering
        \begin{subfigure}{1\textwidth}
        \centering
        \includegraphics[width=1\textwidth]{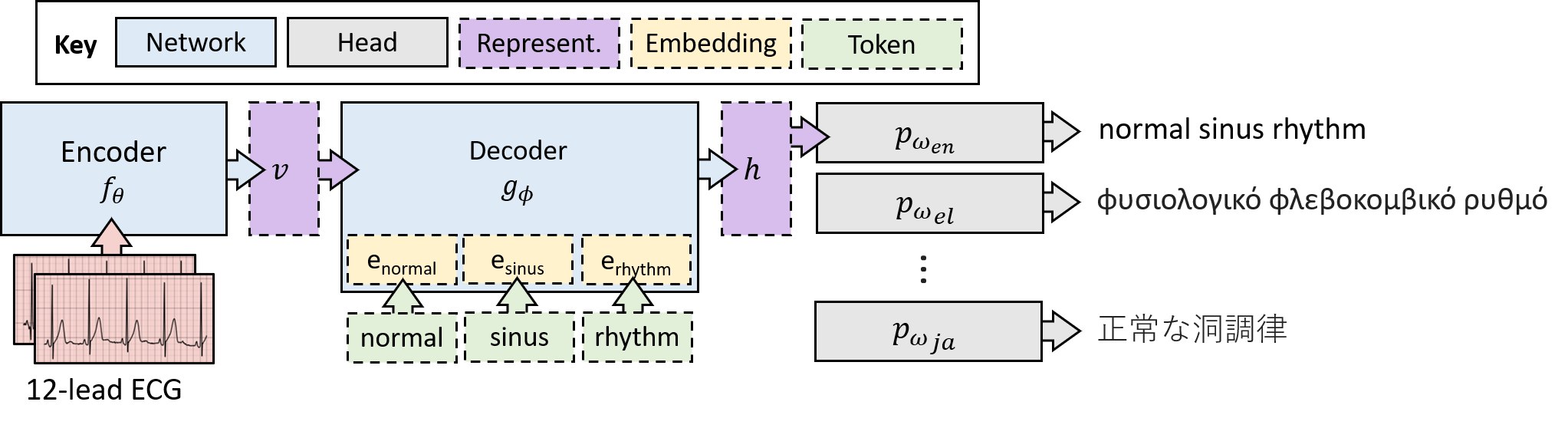}
        \end{subfigure}
    \caption{Multilingual physiological captioning pipeline. A 12-lead ECG is fed into an encoder, $f_{\theta}$, to extract representations, $v$. These are fed, alongside tokens from a vocabulary $V$, of a particular language, to a decoder, $g_{\phi}$, to generate representations, $h$. These representations are then fed into a language-specific MLP, $p_{\omega}$, to generate a caption in a desired language.}
    \label{fig:entire_pipeline}
\end{figure}

We assume access to a set of datasets $\mathcal{D}=\{\mathcal{D}_{l}\}_{l=1}^{N}$, comprising cardiac signals and captions from a set of languages, $l \in \mathrm{L} = \{\mathrm{en},\mathrm{es},\cdots\}$, where $|\mathrm{L}|=N$ and $'\mathrm{en}'$ and $'\mathrm{es}'$ represent English and Spanish captions, respectively. We note that the cardiac signals in this set of datasets are shared. We largely follow the same encoder-decoder approach mentioned in Sec.~\ref{sec:image_captioning}, however we now replace the single MLP head with $N$ MLP heads to account for the distinct vocabularies of the $N$ languages (see Fig.~\ref{fig:entire_pipeline}). This is motivated by recent research demonstrating the importance of a combination of shared and language-specific parameters \citep{Zhang2020MNMT}. More formally, given language-specific parameters, $\omega \in \{\omega_{\mathrm{en}},\omega_{\mathrm{es}}, \cdots \}$, an MLP, $p_{\omega} : h \in \mathbb{R}^{M \times S} \rightarrow y \in \mathbb{R}^{C \times S}$, maps the representations from the decoder, $h$, to a probability distribution over $C$ tokens at each step in the sequence, where $C \in \{|V|_{\mathrm{en}},|V|_{\mathrm{es}}, \cdots \}$ reflects the vocabulary size of a particular language.

As with traditional language models, at each step in the sequence, $s \in [1,S]$, we maximize the likelihood of observing the ground-truth token, $t$, from a particular language, $n \in [1,N]$. Practically, at \textit{each} training iteration, we load $N$ mini-batches each comprising $B$ cardiac signals and captions in one of $N$ languages, and optimize the following multi-task categorical cross-entropy loss. 
\begin{equation}
    \mathcal{L_{\mathrm{multilingual}}} = - \frac{1}{N B S} \sum_{n=1}^{N} \sum_{i=1}^{B} \sum_{s=1}^{S} \log p_{\omega_{n}}(y_{n,i,s} = t_{n,i,s}) 
\end{equation}

\subsection{Discriminative Language Representation Learning}
Modern NLP systems leverage generative pre-training tasks such as masked language modelling \citep{Devlin2018} to learn transferable language representations. More recently, however, \textit{discriminative} pre-training tasks such as ELECTRA \citep{Clark2020} and MARGE \citep{Lewis2020} demonstrated improved transfer to downstream tasks relative to their generative counterparts. This is coupled with the observation that multilingual pre-training yields significant improvements relative to its monolingual counterpart \citep{Conneau2020}. Motivated by the aforementioned findings, we propose a discriminative multilingual pre-training method (described next). 

Before describing our method in depth, we remind readers of our assumption of access to datasets that consist of cardiac signals and multilingual medical reports. To the best of our knowledge, such datasets do not exist. For example, the PTB-XL dataset \citep{Wagner2020} comprises cardiac signals and ECG reports predominantly in German. As a result, we set out to generate such multilingual reports. More specifically, we follow a similar strategy to that proposed by \cite{Conneau2018} and translate reports to multiple languages using the Google Translate API \footnote{\url{https://pypi.org/project/googletrans/}}. Further details on this process can be found in Appendix~\ref{appendix:physiological_captioning_translation}. Although such translation can introduce artifacts, we hypothesize (and indeed show) that the net effect on downstream performance will remain advantageous. We now outline our proposed pre-training method.  

\subsubsection{Replaced Token Language Prediction}
\label{sec:replaced_token_language_detection}
\begin{figure}[!h]
    \centering
        \begin{subfigure}{0.75\textwidth}
        \centering
        \includegraphics[width=1\textwidth]{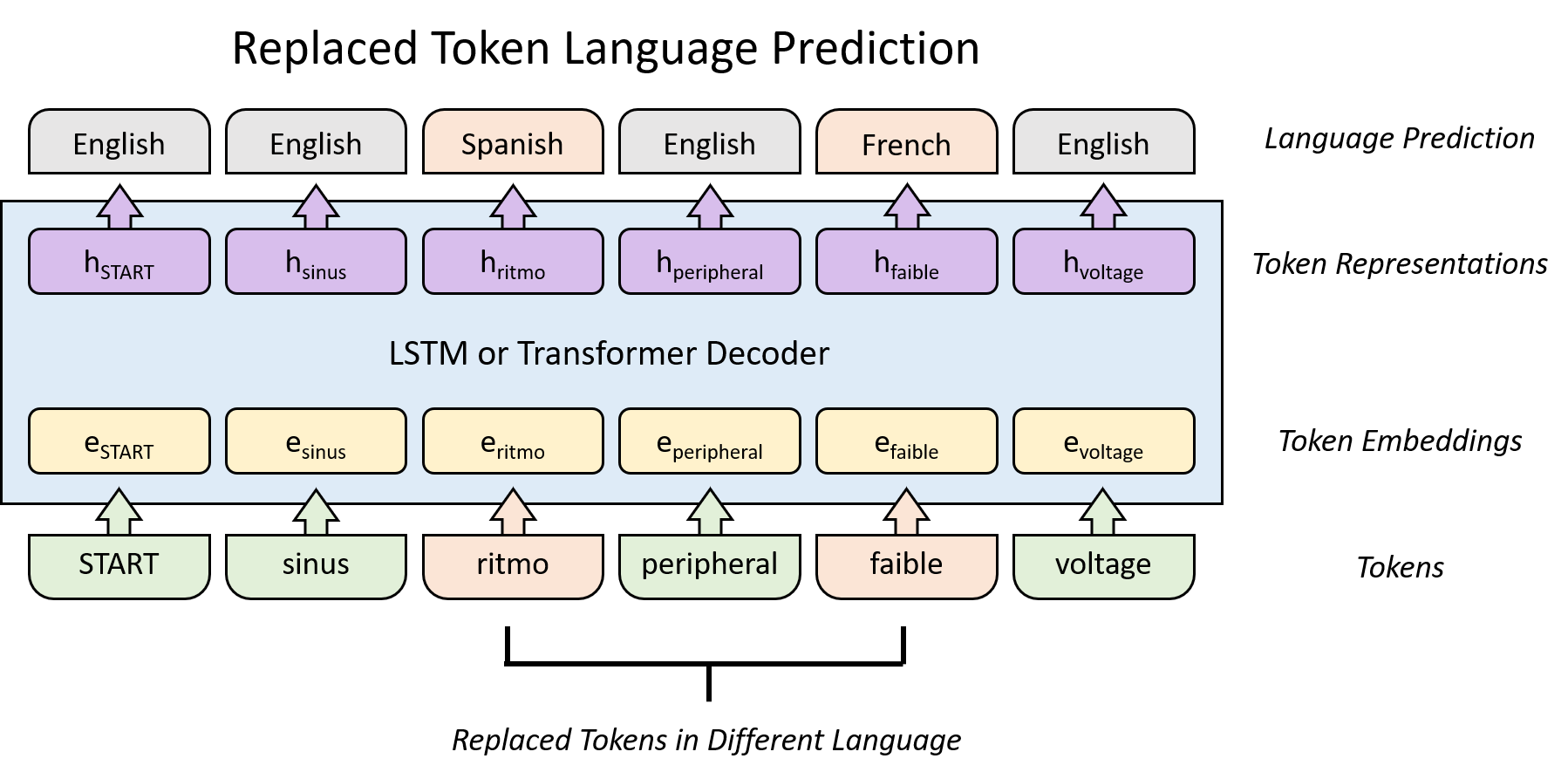}
        \end{subfigure}
    \caption{Proposed discriminative language representation learning method. We sample $k$ locations in the source sentence and replace the corresponding \textit{source} tokens with tokens from a target language. We sample \textit{target} tokens either uniformly at random or from a categorical distribution based on the similarity between the source token embedding and the embeddings of the tokens in the target language. The network is then tasked with predicting the language of each token.}
    \label{fig:decoder_pipeline}
\end{figure}

Multilinguality allows networks to learn representations that transfer well to unseen languages \citep{Dufter2020}. How, then, do we go about imbuing networks with such multilinguality? To model the relationship between medical text from a multitude of languages, we randomly select tokens in a sequence, replace them with tokens from different languages, and task a network with classifying the language of the tokens (see Fig.~\ref{fig:decoder_pipeline}). We now outline this multi-step process in more depth.

\noindent \textbf{Source token selection.} Given a source language, $l_{source} \in \mathrm{L}$, and tokens in a sequence of length, $S$, we randomly sample $k$ distinct locations, $s_{replace} \in [1,S]$, and replace the corresponding source tokens, $x_{source}$, with tokens from a target language, $\mathrm{l}_{target}$. 
\begin{equation}
s_{replace} \sim \mathcal{U}(1,S) \quad |s_{replace}|=k
\end{equation}
\textbf{Target language selection.} We sample a target language uniformly at random from the set of remaining languages $\mathrm{L}_{remain} = \mathrm{L} \setminus l_{source}$ where $|\mathrm{L}_{remain}|=N-1$. 
\begin{equation}
\mathrm{l}_{target} \sim \mathcal{U}(\mathrm{L}_{remain})
\end{equation}
\textbf{Target token selection.} Given a target language's vocabulary, $V_{\mathrm{l}_{target}}$, we can sample a target token, $x_{target}$, uniformly at random (Eq.~\ref{eq:uniform_sampling}). However, this approach may lead to the sampling of a target token that differs semantically from the source token, and thus offer the network a detrimental shortcut to solving the discriminative task of language prediction. To increase the difficulty of the task, we take inspiration from \cite{Dufter2020} and sample a target token which is likely to be a noisy translation of the source token in the target language. We do so by sampling from a categorical distribution, $q$, of the cosine similarity between the source token embedding, $e_{source} \in \mathbb{R}^{M}$, and the embeddings of the tokens, $\left[ e_{j} \in \mathbb{R}^{M} \right]_{j=1}^{|V_{\mathrm{l}_{target}}|}$, in the target language (Eq.~\ref{eq:categorical_sampling}).
\begin{equation}
\label{eq:uniform_sampling}
  x_{target} \sim \mathcal{U}(V_{\mathrm{l}_{target}})
\end{equation}
\begin{align}
\label{eq:categorical_sampling}
   x_{target} &\sim q \quad , \quad \text{where} \quad q_{j} = \frac{\exp^{(e_{source} \cdot e_{j})}}{\sum_{m}^{|V_{\mathrm{l}_{target}}|} \exp^{(e_{source} \cdot e_{m})}} \quad , \quad  \forall x_{j} \in V_{\mathrm{l}_{target}} &
\end{align}

In RTLP, we aim to model the complex interactions between different languages. Given a token (original or replaced) representation, $h$, we define a single classifier head, $p_{\psi}: h \rightarrow y \in \mathbb{R}^{N}$, that predicts the token's associated language. After pre-training, we discard this head. Formally, for $N$ mini-batches each comprising $B$ captions of length, $S$, in one of $N$ source languages, and an indicator function $\delta[r=1]$ where $r=1$ implies that the token was replaced, we optimize the following multi-task categorical cross-entropy loss. 
\begin{equation}
    \mathcal{L_{\mathrm{RTLP}}} = - \frac{1}{N B S} \sum_{n=1}^{N} \sum_{i=1}^{B} \sum_{s=1}^{S} \delta[r=0]\log p_{\psi}(y = n) + \delta[r=1] \log p_{\psi}(y = \text{l}_{target})
\end{equation}

\section{Experimental Design}

\subsection{Datasets}
We focus on a dataset that consists of cardiac signals, such as the electrocardiogram, alongside a paired textual report. To that end, we leverage the publicly-available \textbf{PTB-XL} dataset \citep{Wagner2020} which comprises 12-lead ECG recordings from 18,885 patients alongside ECG reports, which we translate into the following seven languages, German (de), Greek (el), English (en), Spanish (es), French (fr), Italian (it), and Portuguese (pt), and make publicly-available\footnote{\url{https://github.com/danikiyasseh/RTLP}}. The dataset also consists of cardiac arrhythmia labels which we group into 5 major classes \citep{Strodthoff2020}. Further details can be found in Appendix~\ref{appendix:data_description}. 

\subsection{Captioning of Cardiac Signals}

\textbf{Representation learning of cardiac signals.} Neural network pre-training, either supervised or self-supervised, can be effective in learning generalizable representations. To that end, we pre-train the encoder alone in a supervised manner by mapping the 12-lead ECG signals to cardiac arrhythmia labels. Further details of this setup can be found in Appendix~\ref{appendix:physiological_captioning_encoder_pretraining}. We chose this task of cardiac arrhythmia classification because it is well established in the literature and can be directly useful for the downstream task of cardiac signal captioning. The latter can be supported by the observation that ECG reports are heavily contingent upon the presence or absence of cardiac arrhythmia. 

\textbf{Representation learning of language.} In addition to pre-training the encoder, we independently pre-train the decoder and learn language-specific word embeddings. We tokenize the ECG reports by leveraging language-specific tokenizers offered by spaCy\footnote{\url{https://spacy.io/}}, converting the text to lower-case, and removing any punctuation. This results in $N$ distinct vocabularies. Each vocabulary also includes language-specific tokens to indicate the start and end of the report, in addition to the $\mathrm{[PAD]}$ and $\mathrm{[OOV]}$ tokens to refer to padded entries and tokens not seen during training, respectively. We also introduce the $\mathrm{[MASK]}$ token where appropriate. In the case of a Transformer decoder, we feed the representation of each token at the final layer to the MLP. 

Once the encoder and decoder are pre-trained independently of one another, we combine them (as shown in Fig.~\ref{fig:entire_pipeline}) and leverage their learned parameters, in addition to the learned token embeddings, to solve the task of cardiac signal captioning. Using the encoder with frozen parameters, we extract multiple representations per cardiac signal, which we term temporal features. These temporal features can be used to implement either the standard visual attention mechanism \citep{Xu2015} when using an LSTM, or multi-head attention when using a Transformer decoder. We opt for the latter given its empirical superiority. While pre-training and fine-tuning with $N$ languages, each iteration consists of $N$ equally-sized mini-batches of ECG reports covering each of the $N$ languages. This balanced approach is motivated by recent findings which demonstrated inferior performance on languages less frequently-visited \citep{Arivazhagan2019} and thus obviates the need for complex language sampling strategies. Further implementation details can be found in Appendix~\ref{appendix:physiological_captioning_implementation}.

\subsection{Evaluation}
As we are mainly interested in the cardiac captioning task, we leverage metrics commonly used to evaluate image-captioning. These metrics, which include the BLEU score \citep{Papineni2002}, $\mathrm{ROUGE-}\mathrm{L}$ \citep{Lin2004}, and METEOR \citep{Banerjee2005}, predominantly quantify the degree of overlap of n-grams between a ground-truth sentence and a generated sentence.  

\subsection{Baselines}
Our focus is on cardiac signal captioning which exploits multilingual discriminative language representation learning. As such, we compare our method to several language representation learning methods: 1) \textbf{MLM}, a masked language modelling pre-training objective \citep{Devlin2018} where the decoder is tasked with identifying masked tokens, 2) \textbf{ELECTRA}, a replaced token detection pre-training objective \citep{Clark2020} where the decoder is tasked with identifying whether tokens have been replaced with those from an MLM model, and 3) \textbf{MARGE}, a multilingual generative language representation learning approach \citep{Lewis2020} where source documents in various languages are exploited to generate a similar yet distinct target document. Further details on how we adapted these methods can be found in Appendix~\ref{appendix:physiological_captioning_baseline_implementations}. 

\subsection{Hyperparameters}
We conduct our experiments\footnote{Our code is available at \url{https://github.com/danikiyasseh/RTLP}} using PyTorch \citep{Paszke2019} and the Adam optimizer. We pre-train the encoder and decoder with a patience value of 10 and 25 epochs, respectively, on the validation loss. When transferring the parameters to the task of cardiac signal captioning, we use those associated with the lowest validation loss. When fine-tuning, we checkpoint the parameters associated with the highest validation BLEU score.

\section{Experimental Results}

\subsection{Multilingual Captioning of Cardiac Signals}
\label{sec:captioning_results}
We are primarily interested in the utility of multilingual language representation learning in the context of multilingual cardiac signal captioning. In this section, we quantify this utility as it pertains to our proposed methods and several state-of-the-art approaches. In Table~\ref{table:physio_captioning_bleu_ptbxl}, we illustrate the $\mathrm{BLEU-}1$, METEOR, and $\mathrm{ROUGE-}\mathrm{L}$ scores of the cardiac signal captioning task when evaluated in seven different languages on the PTB-XL dataset.  

\begin{table}[!h]
    \centering
    \caption{Evaluation of the various pre-training methods when captioning cardiac signals in the validation set of PTB-XL across five seeds. The standard deviation is shown in brackets. We find that RTLP outperforms the state-of-the-art discriminative language pre-training method, ELECTRA, and performs on par with more expensive state-of-the-art pre-training methods, MLM and MARGE.}
    \label{table:physio_captioning_bleu_ptbxl}
    \begin{adjustbox}{max width=\textwidth}
    \begin{tabular}{c | c c c c c c c | c}
    \toprule
         \multirow{1}{*}{\textbf{Method}} & \textbf{de} & \textbf{el} & \textbf{en} & \textbf{es} & \textbf{fr} & \textbf{it} & \textbf{pt} & \textbf{avg} \\
    \midrule
    \multicolumn{9}{l}{\textit{BLEU-1}} \\
    \midrule
 MLM & 25.9 \scriptsize(0.6) & 20.5 \scriptsize(0.3) & 31.3 \scriptsize(0.5) & 33.2 \scriptsize(0.8) & 29.7 \scriptsize(0.6) & 30.3 \scriptsize(0.2) & 34.9 \scriptsize(0.7) & 29.4 \scriptsize(4.6) \\
 ELECTRA & 0.1 \scriptsize(0.1) & 0.2 & 0.2 \scriptsize(0.2) & 0.3 \scriptsize(0.1) & 0.6 \scriptsize(0.1) & 0.5 \scriptsize(0.1) & 0.5 \scriptsize(0.1) & 0.3 \scriptsize(0.2) \\
 MARGE & 24.9 \scriptsize(1.0) & 19.5 \scriptsize(0.9) & 30.8 \scriptsize(0.5) & 32.9 \scriptsize(0.5) & 29.7 \scriptsize(0.6) & 29.4 \scriptsize(0.5) & 34.5 \scriptsize(1.0) & 28.9 \scriptsize(4.8)\\
 RTLP (eq.~\ref{eq:uniform_sampling}) & 25.4 \scriptsize(1.3) & 20.5 \scriptsize(0.5) & 31.0 \scriptsize(0.4) & 33.0 \scriptsize(0.6) & 29.8 \scriptsize(0.2) & 30.5 \scriptsize(0.5) & 34.7 \scriptsize(1.2) & 29.3 \scriptsize(4.6) \\
 RTLP (eq.~\ref{eq:categorical_sampling}) & 25.4 \scriptsize(1.1) & 19.8 \scriptsize(0.6) & 30.0 \scriptsize(0.7) & 33.1 \scriptsize(0.9) & 28.3 \scriptsize(0.8) & 30.0 \scriptsize(0.1) & 33.5 \scriptsize(1.0) & 28.5 \scriptsize(4.5) \\
    \midrule
    \multicolumn{9}{l}{\textit{METEOR}} \\
    \midrule
 MLM & 36.7 \scriptsize(1.0) & 23.6 \scriptsize(0.2) & 37.3 \scriptsize(1.1) & 38.6 \scriptsize(0.6) & 33.5 \scriptsize(0.7) & 33.9 \scriptsize(0.7) & 38.8 \scriptsize(0.7) & 34.6 \scriptsize(5.0) \\
 ELECTRA & 0.3 \scriptsize(0.5) & 0.2 \scriptsize(0.1) & 0.2 \scriptsize(0.1) & 0.5 \scriptsize(0.4) & 1.1 \scriptsize(0.3) & 0.9 \scriptsize(0.2) & 0.5 \scriptsize(0.2) & 0.5 \scriptsize(0.4)\\
 MARGE & 35.6 \scriptsize(1.5) & 22.2 \scriptsize(1.0) & 36.5 \scriptsize(0.8) & 37.1 \scriptsize(0.6) & 33.1 \scriptsize(1.0) & 32.9 \scriptsize(0.9) & 37.8 \scriptsize(1.1) & 33.6 \scriptsize(5.1) \\
 RTLP (eq.~\ref{eq:uniform_sampling}) & 35.6 \scriptsize(1.6) & 23.5 \scriptsize(0.5) & 36.8 \scriptsize(0.3) & 38.5 \scriptsize(0.5) & 33.3 \scriptsize(0.2) & 34.0 \scriptsize(0.7) & 38.7 \scriptsize(0.8) & 34.3 \scriptsize(5.0) \\
 RTLP (eq.~\ref{eq:categorical_sampling}) & 36.5 \scriptsize(0.7) & 22.6 \scriptsize(1.0) & 36.0 \scriptsize(0.9) & 38.5 \scriptsize(0.8) & 32.4 \scriptsize(0.4) & 33.7 \scriptsize(0.9) & 37.6 \scriptsize(0.6) & 33.9 \scriptsize(5.1) \\
    \midrule
    \multicolumn{9}{l}{\textit{ROUGE-L}} \\
    \midrule
 MLM & 34.6 \scriptsize(0.8) & 11.4 \scriptsize(1.9) & 28.5 \scriptsize(0.2) & 39.3 \scriptsize(1.3) & 34.5 \scriptsize(0.8) & 36.9 \scriptsize(0.5) & 39.1 \scriptsize(0.6) & 33.5 \scriptsize(9.3) \\
 ELECTRA & 0.2 \scriptsize(0.3) & 0 & 0.2 & 0.5 \scriptsize(0.3) & 1.0 \scriptsize(0.2) & 0.8 \scriptsize(0.1) & 0.5 \scriptsize(0.1) & 0.5 \scriptsize(0.4) \\
 MARGE & 33.2 \scriptsize(0.7) & 11.1 \scriptsize(2.3) & 38.1 \scriptsize(0.5) & 39.2 \scriptsize(0.6) & 34.4 \scriptsize(0.8) & 36.1 \scriptsize(0.6) & 39.0 \scriptsize(0.5) & 33.0 \scriptsize(9.4) \\
 RTLP (eq.~\ref{eq:uniform_sampling}) & 34.6 \scriptsize(0.8) & 10.6 \scriptsize(2.0) & 38.4 \scriptsize(0.5) & 39.4 \scriptsize(0.8) & 34.8 \scriptsize(0.3) & 37.4 \scriptsize(0.7) & 38.9 \scriptsize(1.0) & 33.4 \scriptsize(9.7) \\
  RTLP (eq.~\ref{eq:categorical_sampling}) & 34.0 \scriptsize(1.1) & 11.6 \scriptsize(2.3) & 36.3 \scriptsize(0.9) & 39.1 \scriptsize(1.2) & 33.1 \scriptsize(0.9) & 36.5 \scriptsize(1.0) & 37.3 \scriptsize(0.9) & 32.6 \scriptsize(8.9) \\
    \bottomrule
    \end{tabular}
    \end{adjustbox}
\end{table}

In Table~\ref{table:physio_captioning_bleu_ptbxl}, we show that RTLP performs on par with state-of-the-art generative pre-training methods (MLM, MARGE) that are typically quite expensive to implement. For example, RTLP achieves $\mathrm{BLEU-}1=29.3$ whereas MLM and MARGE achieve $\mathrm{BLEU-}1=29.4$ and $28.9$, respectively. This finding holds regardless of the language in which the reports are generated and the evaluation metric that is used. Namely, we arrive at a similar conclusion when inspecting the METEOR and $\mathrm{ROUGE-}\mathrm{L}$ scores. Furthermore, we find that RTLP outperforms the state-of-the-art discriminative pre-training method, ELECTRA, by a significant margin. For example, RTLP achieves $\mathrm{ROUGE-}\mathrm{L}=33.4$ whereas ELECTRA achieves $\mathrm{ROUGE-}\mathrm{L}=0.5$. The aforementioned findings suggest that RTLP can be an effective discriminative pre-training method and a potential alternative to existing generative methods. We also find that, regardless of the method implemented, performance can vary significantly across languages. For example, MLM achieves $\mathrm{BLEU-}1=20.5$ and $31.3$ on Greek (el) and English (en) reports, respectively. We hypothesize that the poorer performance on the Greek language is due to its relatively unique alphabet in our pool of languages, which makes it less likely to benefit from knowledge shared across languages during training. 

\subsection{Inspecting Generated Multilingual Reports}
\label{sec:captioning}
So far, we have shown that a network pre-trained with RTLP is capable of performing on par with networks pre-trained with more expensive generative methods. Beyond achieving strong generalization performance, however, we aim to generate plausible clinical text. To determine whether we have been successful in achieving this, we illustrate, in Table~\ref{tab:signals_and_tokens}, 12-Lead ECG segments alongside the target and generated multilingual ECG reports.

\begin{table}[!h]
\captionof{table}{12-lead ECG segments alongside the multilingual target reports and those generated by the network pre-trained with the RTLP paradigm. Note that punctuation was removed as part of the pre-processing of the reports. We show that the system generates relatively accurate and diverse clinical reports in multiple languages.}
\label{tab:signals_and_tokens}
\begin{minipage}{1\textwidth}
\centering
\includegraphics[width=0.7\textwidth]{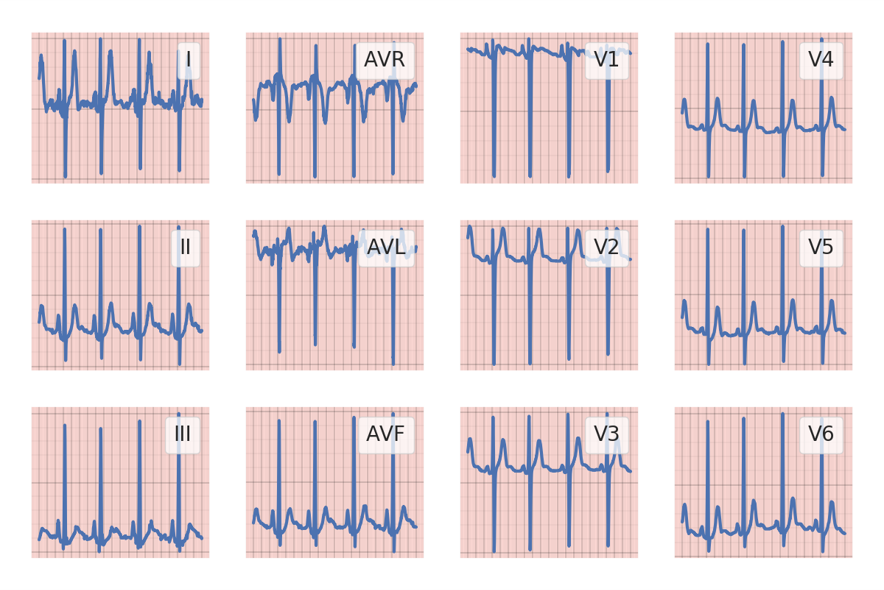}
\end{minipage}\hfill

\begin{minipage}{1\textwidth}
\centering
\scriptsize
\begin{tabular}{c p{0.40\textwidth} p{0.40\textwidth} }
\toprule
Lang & Target Report & Generated Report \\
\midrule
\rowcolor{tableblue}
\textbf{de} & rhythm st segments are depressed in i avl v6. t waves are inverted in i avl v5,6. this may be due to lv strain or ischaemia & sinus rhythm t waves are flat in i avl v5,6. this may be due to lv strain or ischaemia\\
\textbf{el} & \textgreek{ρυθμό τα τμήματα είναι πιεσμένα στο i avl v6 τα κύματα t αντιστρέφονται σε i avl v5,6 αυτό μπορεί να οφείλεται σε στέλεχος lv ή ισχαιμία} & \textgreek{φλεβοκομβικό ρυθμό τα τμήματα πιέζονται στο i avl τα κύματα t είναι χαμηλά ή επίπεδα σε i avl v5,6 αυτό μπορεί να οφείλεται σε στέλεχος lv ή ισχαιμία} \\
\rowcolor{tableblue}
\textbf{en} & rhythm st segments are depressed in i avl v6 t waves are inverted in i avl v5,6 this may be due to lv strain or ischaemia & sinus rhythm location type normal qrs t abnormal inferior myocardial damage possible t abnormal in anterolateral leads 4.46 unconfirmed report\\
\textbf{es} & sinusal los segmentos st están deprimidos en i avl v6 t ondas están invertidas en i avl v5,6 esto puede deberse a una tensión lv o isquemia & ritmo sinusal tipo de localización normal inespecífico anormal t 4,46 informe no confirmado \\
\rowcolor{tableblue}
\textbf{fr }& rythme sinusal \underline{st segments sont enfoncés} dans i avl v6 les ondes \underline{t sont inversées} dans i avl v5,6 cela peut être dû à une souche lv ou à une ischémie & un rythme sinusal \underline{st segments sont enfoncés} dans i avl les ondes \underline{t sont inversées} dans i avl et flat dans v5,6 les résultats sont probablement dus à une cardiopathie ischémique l' âge des changements est incertain \\
\textbf{it} & sinusale i segmenti st sono depressi in i avl v6 le onde t sono invertite in i avl v5,6 ciò può essere dovuto a ceppo lv o ischemia & ritmo sinusale i segmenti st sono depressi in i avl le onde t sono piatte in i avl v6 ciò può essere dovuto a ceppo lv o ischemia \\
\rowcolor{tableblue}
\textbf{pt} & sinusal os segmentos st estão deprimidos em i avl v6 ondas t são invertidas em i avl v5,6 isso pode ser devido à cepa lv ou isquemia & ritmo sinusal os segmentos st estão deprimidos em i avl ondas t são invertidas em i avl v5,6 isso pode ser devido à cepa lv ou isquemia \\
\bottomrule
\end{tabular}
\end{minipage}

\end{table}

In Table~\ref{tab:signals_and_tokens}, we find that the generated ECG reports are indeed plausible and manage to capture the general pathology reflected in the target report without simply regurgitating content. For example, when inspecting the generated French (fr) report, we see that the phrases \textit{\enquote{ST segments sont enfonc\'es}} (ST segments are depressed) and \textit{\enquote{T sont invers\'ees}} (T waves are inverted) are appropriately included and refer to a cardiac pathology. More interestingly, the cause of this pathology was expressed differently, albeit correctly, to that shown in the target report. Whereas the target report states \textit{\enquote{peut etre \ldots une ischemie}} to indicate that an ischemia is the likely culprit, the generated report states \textit{\enquote{les r\'esultats sont probablement dus a une cardiopathie isch\'emique}}, echoing the same idea in a different manner. Such a finding suggests that our model is capable of not only capturing the high-level medical condition reflected in the ECG but also generating diverse text; that which extends beyond naive regurgitation. 

\subsection{Quantifying Diversity of Generated Multilingual Reports}
\begin{figure}[!h]
    \centering
        \begin{subfigure}{\textwidth}
        \centering
        \includegraphics[width=0.7\textwidth]{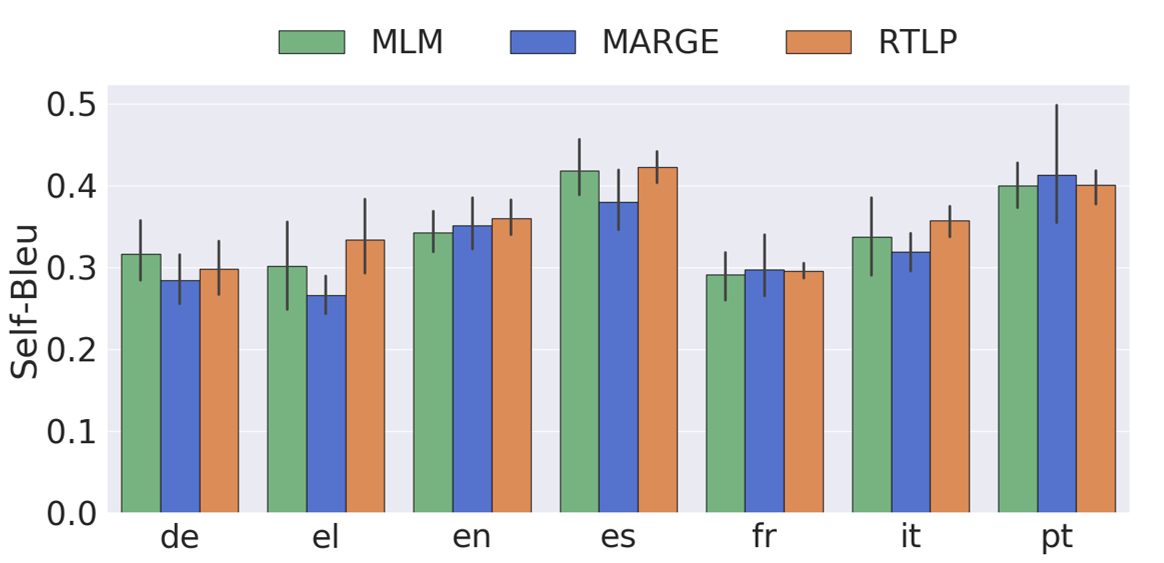}
        \end{subfigure}
    \caption{Self-BLEU of the clinical text, in each of the six languages, generated by the multilingual MLM, MARGE, and RTLP (ours) models. $\uparrow$ Self-BLEU implies increased similarity between sentences and thus $\downarrow$ diversity. Error bars indicate the standard deviation across five seeds.}
    \label{fig:self_bleu_text}
\end{figure}

It could be argued that a desirable captioning system is one that generates both plausible and \textit{diverse} text. However, it has been documented that the diversity of generated text can be at odds with the performance of captioning systems \citep{Duvsek2020,Tevet2020}. In previous sections, we made the case for plausibility. In this section, we shift our emphasis toward quantifying the diversity of generated text, and do so via the Self-BLEU metric proposed by \cite{Zhu2018}. The lower the value of this metric ($\downarrow$ Self-BLEU), which measures the BLEU score between all pairs of generated reports, the more diverse the reports are ($\uparrow$ diversity). In Fig.~\ref{fig:self_bleu_text}, we illustrate the Self-BLEU score for three different pre-training methods across all languages in the multilingual setting. 

There are several takeaways from Fig.~\ref{fig:self_bleu_text}. First, the degree of diversity of the generated reports can vary significantly across languages. For example, we find that all three models achieve Self-BLEU$\approx 0.30$ on French (fr) reports however a Self-BLEU$\approx 0.40$ on Spanish (es) reports. Second, we find that the degree of diversity of the generated reports is similar across the three models, $\mathrm{MLM}$, $\mathrm{MARGE}$, and $\mathrm{RTLP}$. This statement holds regardless of the language in which the report is being generated. Such a finding, coupled with the results in Secs.~\ref{sec:captioning_results} and \ref{sec:captioning}, adds further evidence in support of our pre-training method as a reliable alternative to existing generative pre-training methods. 

\subsection{Investigating the Curse of Multilinguality}
Having established that networks pre-trained with RTLP are capable of generating relatively reliable and plausible clinical text, we wanted to explore whether our framework was experiencing symptoms associated with the \enquote{curse of multilinguality} \citep{Conneau2019}. Concisely, this curse attributes the potentially poorer performance of multilingual models relative to their monolingual counterparts to interference between the various languages. This is conceptually similar to interference that may be experienced during multi-task learning \citep{Caruana1993}. 

\begin{figure}[!h]
    \centering
    
    \begin{subfigure}{0.48\textwidth}
        \centering
        \includegraphics[width=1\textwidth]{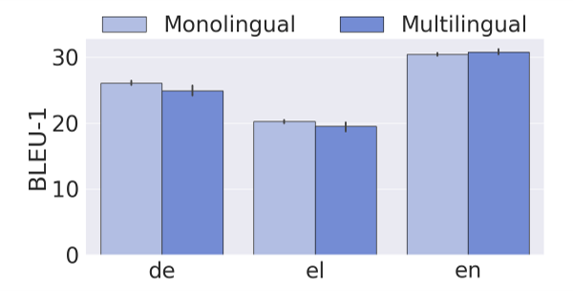}
        \caption{MARGE}
        \label{fig:MARGE_curse_multilinguality}
    \end{subfigure}
    \begin{subfigure}{0.48\textwidth}
        \centering
        \includegraphics[width=1\textwidth]{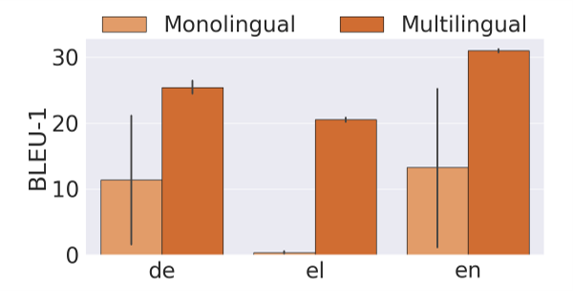}
        \caption{RTLP}
        \label{fig:RTLP_curse_multilinguality}
    \end{subfigure}
    
    \caption{Performance of cardiac signal captioning systems when pre-trained with (a) MARGE or (b) RTLP, and fine-tuned in a monolingual and multilingual setting. The multilingual setting involves all seven languages [de,el,en,es,fr,it,pt]. In (a), the curse of multilinguality is almost non-existent. In (b), the blessing of multilinguality is evident.}
    \label{fig:curse_multilinguality}
\end{figure}

To determine the presence of the \enquote{curse of multilinguality} and quantify its effect on the performance of our networks, we conduct fine-tuning experiments in the monolingual setting. In other words, we task our network with generating captions in a single language. In Fig.~\ref{fig:curse_multilinguality}, we illustrate the $\mathrm{BLEU-}1$ score of various cardiac signal captioning systems when fine-tuned in the monolingual and multilingual setting. 

In Fig.~\ref{fig:MARGE_curse_multilinguality}, we find that the curse of multilinguality is almost non-existent. This can be seen by the similar performance exhibited by the networks fine-tuned in the monolingual and multilingual setting. For example, when generating reports in German (de), both networks achieve $\mathrm{BLEU-}1\approx 25$. This finding holds across languages and evaluation metrics (not shown). In Fig.~\ref{fig:RTLP_curse_multilinguality}, we find that the multilingual setting is more advantageous than its monolingual counterpart. In contrast to the detrimental curse of multilinguality, we denote this beneficial finding as the \enquote{blessing of multilinguality}. We show that networks fine-tuned in the multilingual setting outperform those fine-tuned in the monolingual setting. For example, when generating reports in German (de), the two networks achieve $\mathrm{BLEU-}1\approx 0.25$ and $\approx 0.11$, respectively. Such a finding suggests that, during multilingual fine-tuning, knowledge is being shared across languages in a beneficial way. 

\section{Discussion and Future Work}

In this paper, we introduced a novel task, which we refer to as multilingual captioning of cardiac signals, in which a neural network receives a cardiac signal as input and returns a clinical report in multiple languages as output. To warm-start the language decoder, we proposed a discriminative multilingual pre-training paradigm, entitled \textit{replaced token language prediction} (RTLP), where tokens are randomly replaced with those from different languages and a neural network is tasked with predicting the language of all tokens. We showed that RTLP performs on par with state-of-the-art generative and discriminative pre-training methods such as MLM, ELECTRA, and MARGE. We also demonstrated that our method generates diverse and plausible clinical text that mirrors the diversity of text generated by baseline methods. Lastly, we showed that multilingual models can sometimes outperform their monolingual counterparts, shedding light on the \enquote{blessing of multilinguality}. We now elucidate several avenues worth exploring.

\textbf{Complexity of clinical reports.} We have focused on the generation of ECG reports in response to cardiac signals. However, the text contained within ECG reports can be limited in diversity. Extending our work to reports from different clinical domains that exhibit increased complexity will reaffirm the added value of our method.

\textbf{Multi-modal cardiac signal captioning.} Clinicians typically leverage multiple data modalities to arrive at a clinical decision. Therefore, a medical captioning system stands to benefit from multiple modalities, e.g., ECG, coronary angiograms, etc., in order to generate a more holistic clinical report.

\bibliography{iclr2021_conference}

\begin{thebibliography}{49}
\providecommand{\natexlab}[1]{#1}
\providecommand{\url}[1]{\texttt{#1}}
\expandafter\ifx\csname urlstyle\endcsname\relax
  \providecommand{\doi}[1]{doi: #1}\else
  \providecommand{\doi}{doi: \begingroup \urlstyle{rm}\Url}\fi

\bibitem[Anderson et~al.(2018)Anderson, He, Buehler, Teney, Johnson, Gould, and
  Zhang]{Anderson2018}
Peter Anderson, Xiaodong He, Chris Buehler, Damien Teney, Mark Johnson, Stephen
  Gould, and Lei Zhang.
\newblock Bottom-up and top-down attention for image captioning and visual
  question answering.
\newblock In \emph{Proceedings of the IEEE conference on computer vision and
  pattern recognition}, pp.\  6077--6086, 2018.

\bibitem[Arivazhagan et~al.(2019)Arivazhagan, Bapna, Firat, Lepikhin, Johnson,
  Krikun, Chen, Cao, Foster, Cherry, et~al.]{Arivazhagan2019}
Naveen Arivazhagan, Ankur Bapna, Orhan Firat, Dmitry Lepikhin, Melvin Johnson,
  Maxim Krikun, Mia~Xu Chen, Yuan Cao, George Foster, Colin Cherry, et~al.
\newblock Massively multilingual neural machine translation in the wild:
  Findings and challenges.
\newblock \emph{arXiv preprint arXiv:1907.05019}, 2019.

\bibitem[Artetxe et~al.(2020)Artetxe, Labaka, and Agirre]{Artetxe2020}
Mikel Artetxe, Gorka Labaka, and Eneko Agirre.
\newblock Translation artifacts in cross-lingual transfer learning.
\newblock \emph{arXiv preprint arXiv:2004.04721}, 2020.

\bibitem[Banerjee \& Lavie(2005)Banerjee and Lavie]{Banerjee2005}
Satanjeev Banerjee and Alon Lavie.
\newblock Meteor: An automatic metric for mt evaluation with improved
  correlation with human judgments.
\newblock In \emph{Proceedings of the acl workshop on intrinsic and extrinsic
  evaluation measures for machine translation and/or summarization}, pp.\
  65--72, 2005.

\bibitem[Brailer et~al.(1997)Brailer, Kroch, and Pauly]{Brailer1997}
David~J Brailer, Eugene Kroch, and Mark~V Pauly.
\newblock The impact of computer-assisted test interpretation on physician
  decision making: the case of electrocardiograms.
\newblock \emph{Medical decision making}, 17\penalty0 (1):\penalty0 80--86,
  1997.

\bibitem[Caruana(1993)]{Caruana1993}
Richard~A Caruana.
\newblock Multitask connectionist learning.
\newblock In \emph{In Proceedings of the 1993 Connectionist Models Summer
  School}. Citeseer, 1993.

\bibitem[Clark et~al.(2020)Clark, Luong, Le, and Manning]{Clark2020}
Kevin Clark, Minh-Thang Luong, Quoc~V Le, and Christopher~D Manning.
\newblock Electra: Pre-training text encoders as discriminators rather than
  generators.
\newblock \emph{arXiv preprint arXiv:2003.10555}, 2020.

\bibitem[Conneau \& Lample(2019)Conneau and Lample]{Conneau2019NIPS}
Alexis Conneau and Guillaume Lample.
\newblock Cross-lingual language model pretraining.
\newblock In \emph{Advances in Neural Information Processing Systems}, pp.\
  7059--7069, 2019.

\bibitem[Conneau et~al.(2018)Conneau, Lample, Rinott, Williams, Bowman,
  Schwenk, and Stoyanov]{Conneau2018}
Alexis Conneau, Guillaume Lample, Ruty Rinott, Adina Williams, Samuel~R Bowman,
  Holger Schwenk, and Veselin Stoyanov.
\newblock Xnli: Evaluating cross-lingual sentence representations.
\newblock \emph{arXiv preprint arXiv:1809.05053}, 2018.

\bibitem[Conneau et~al.(2019)Conneau, Khandelwal, Goyal, Chaudhary, Wenzek,
  Guzm{\'a}n, Grave, Ott, Zettlemoyer, and Stoyanov]{Conneau2019}
Alexis Conneau, Kartikay Khandelwal, Naman Goyal, Vishrav Chaudhary, Guillaume
  Wenzek, Francisco Guzm{\'a}n, Edouard Grave, Myle Ott, Luke Zettlemoyer, and
  Veselin Stoyanov.
\newblock Unsupervised cross-lingual representation learning at scale.
\newblock \emph{arXiv preprint arXiv:1911.02116}, 2019.

\bibitem[Conneau et~al.(2020)Conneau, Baevski, Collobert, Mohamed, and
  Auli]{Conneau2020}
Alexis Conneau, Alexei Baevski, Ronan Collobert, Abdelrahman Mohamed, and
  Michael Auli.
\newblock Unsupervised cross-lingual representation learning for speech
  recognition.
\newblock \emph{arXiv preprint arXiv:2006.13979}, 2020.

\bibitem[Devlin et~al.(2018)Devlin, Chang, Lee, and Toutanova]{Devlin2018}
Jacob Devlin, Ming-Wei Chang, Kenton Lee, and Kristina Toutanova.
\newblock Bert: Pre-training of deep bidirectional transformers for language
  understanding.
\newblock \emph{arXiv preprint arXiv:1810.04805}, 2018.

\bibitem[Dufter \& Sch{\"u}tze(2020)Dufter and Sch{\"u}tze]{Dufter2020}
Philipp Dufter and Hinrich Sch{\"u}tze.
\newblock Identifying elements essential for bert’s multilinguality.
\newblock In \emph{Proceedings of the 2020 Conference on Empirical Methods in
  Natural Language Processing (EMNLP)}, pp.\  4423--4437, 2020.

\bibitem[Du{\v{s}}ek et~al.(2020)Du{\v{s}}ek, Novikova, and Rieser]{Duvsek2020}
Ond{\v{r}}ej Du{\v{s}}ek, Jekaterina Novikova, and Verena Rieser.
\newblock Evaluating the state-of-the-art of end-to-end natural language
  generation: The e2e nlg challenge.
\newblock \emph{Computer Speech \& Language}, 59:\penalty0 123--156, 2020.

\bibitem[Goyal et~al.(2017)Goyal, Khot, Summers-Stay, Batra, and
  Parikh]{Goyal2017}
Yash Goyal, Tejas Khot, Douglas Summers-Stay, Dhruv Batra, and Devi Parikh.
\newblock Making the v in vqa matter: Elevating the role of image understanding
  in visual question answering.
\newblock In \emph{Proceedings of the IEEE Conference on Computer Vision and
  Pattern Recognition}, pp.\  6904--6913, 2017.

\bibitem[Hasan et~al.(2018)Hasan, Ling, Liu, Sreenivasan, Anand, Arora, Datla,
  Lee, Qadir, Swisher, et~al.]{Hasan2018}
Sadid~A Hasan, Yuan Ling, Joey Liu, Rithesh Sreenivasan, Shreya Anand,
  Tilak~Raj Arora, Vivek Datla, Kathy Lee, Ashequl Qadir, Christine Swisher,
  et~al.
\newblock Attention-based medical caption generation with image modality
  classification and clinical concept mapping.
\newblock In \emph{International Conference of the Cross-Language Evaluation
  Forum for European Languages}, pp.\  224--230. Springer, 2018.

\bibitem[Herdade et~al.(2019)Herdade, Kappeler, Boakye, and
  Soares]{Herdade2019}
Simao Herdade, Armin Kappeler, Kofi Boakye, and Joao Soares.
\newblock Image captioning: Transforming objects into words.
\newblock In \emph{Advances in Neural Information Processing Systems}, pp.\
  11137--11147, 2019.

\bibitem[Hibbard et~al.(2001)Hibbard, Peters, Slovic, Finucane, and
  Tusler]{Hibbard2001}
Judith~H Hibbard, Ellen Peters, Paul Slovic, Melissa~L Finucane, and Martin
  Tusler.
\newblock Making health care quality reports easier to use.
\newblock \emph{The Joint Commission journal on quality improvement},
  27\penalty0 (11):\penalty0 591--604, 2001.

\bibitem[Huang et~al.(2019{\natexlab{a}})Huang, Liang, Duan, Gong, Shou, Jiang,
  and Zhou]{Huang2019Unicoder}
Haoyang Huang, Yaobo Liang, Nan Duan, Ming Gong, Linjun Shou, Daxin Jiang, and
  Ming Zhou.
\newblock Unicoder: A universal language encoder by pre-training with multiple
  cross-lingual tasks.
\newblock \emph{arXiv preprint arXiv:1909.00964}, 2019{\natexlab{a}}.

\bibitem[Huang et~al.(2019{\natexlab{b}})Huang, Altosaar, and
  Ranganath]{Huang2019}
Kexin Huang, Jaan Altosaar, and Rajesh Ranganath.
\newblock Clinicalbert: Modeling clinical notes and predicting hospital
  readmission.
\newblock \emph{arXiv preprint arXiv:1904.05342}, 2019{\natexlab{b}}.

\bibitem[Jin et~al.(2019)Jin, Dhingra, Cohen, and Lu]{Jin2019}
Qiao Jin, Bhuwan Dhingra, William~W Cohen, and Xinghua Lu.
\newblock Probing biomedical embeddings from language models.
\newblock \emph{arXiv preprint arXiv:1904.02181}, 2019.

\bibitem[Keselman \& Smith(2012)Keselman and Smith]{Keselman2012}
Alla Keselman and Catherine~Arnott Smith.
\newblock A classification of errors in lay comprehension of medical documents.
\newblock \emph{Journal of biomedical informatics}, 45\penalty0 (6):\penalty0
  1151--1163, 2012.

\bibitem[Kisilev et~al.(2016)Kisilev, Sason, Barkan, and Hashoul]{Kisilev2016}
Pavel Kisilev, Eli Sason, Ella Barkan, and Sharbell Hashoul.
\newblock Medical image captioning: Learning to describe medical image findings
  using multi-task-loss cnn.
\newblock \emph{Deep Learning for Precision Medicine, Riva del Garda, Italy},
  2016.

\bibitem[Lee et~al.(2020)Lee, Yoon, Kim, Kim, Kim, So, and Kang]{Lee2020}
Jinhyuk Lee, Wonjin Yoon, Sungdong Kim, Donghyeon Kim, Sunkyu Kim, Chan~Ho So,
  and Jaewoo Kang.
\newblock Biobert: a pre-trained biomedical language representation model for
  biomedical text mining.
\newblock \emph{Bioinformatics}, 36\penalty0 (4):\penalty0 1234--1240, 2020.

\bibitem[Lewis et~al.(2020)Lewis, Ghazvininejad, Ghosh, Aghajanyan, Wang, and
  Zettlemoyer]{Lewis2020}
Mike Lewis, Marjan Ghazvininejad, Gargi Ghosh, Armen Aghajanyan, Sida Wang, and
  Luke Zettlemoyer.
\newblock Pre-training via paraphrasing.
\newblock \emph{arXiv preprint arXiv:2006.15020}, 2020.

\bibitem[Lin(2004)]{Lin2004}
Chin-Yew Lin.
\newblock Rouge: A package for automatic evaluation of summaries.
\newblock In \emph{Text summarization branches out}, pp.\  74--81, 2004.

\bibitem[Liu et~al.(2019{\natexlab{a}})Liu, Hsu, McDermott, Boag, Weng,
  Szolovits, and Ghassemi]{Liu2019}
Guanxiong Liu, Tzu-Ming~Harry Hsu, Matthew McDermott, Willie Boag, Wei-Hung
  Weng, Peter Szolovits, and Marzyeh Ghassemi.
\newblock Clinically accurate chest x-ray report generation.
\newblock \emph{arXiv preprint arXiv:1904.02633}, 2019{\natexlab{a}}.

\bibitem[Liu et~al.(2019{\natexlab{b}})Liu, Ott, Goyal, Du, Joshi, Chen, Levy,
  Lewis, Zettlemoyer, and Stoyanov]{Liu2019Roberta}
Yinhan Liu, Myle Ott, Naman Goyal, Jingfei Du, Mandar Joshi, Danqi Chen, Omer
  Levy, Mike Lewis, Luke Zettlemoyer, and Veselin Stoyanov.
\newblock Roberta: A robustly optimized bert pretraining approach.
\newblock \emph{arXiv preprint arXiv:1907.11692}, 2019{\natexlab{b}}.

\bibitem[Liu et~al.(2020)Liu, Gu, Goyal, Li, Edunov, Ghazvininejad, Lewis, and
  Zettlemoyer]{Liu2020}
Yinhan Liu, Jiatao Gu, Naman Goyal, Xian Li, Sergey Edunov, Marjan
  Ghazvininejad, Mike Lewis, and Luke Zettlemoyer.
\newblock Multilingual denoising pre-training for neural machine translation.
\newblock \emph{arXiv preprint arXiv:2001.08210}, 2020.

\bibitem[Lu et~al.(2019)Lu, Batra, Parikh, and Lee]{Lu2019}
Jiasen Lu, Dhruv Batra, Devi Parikh, and Stefan Lee.
\newblock Vilbert: Pretraining task-agnostic visiolinguistic representations
  for vision-and-language tasks.
\newblock In \emph{Advances in Neural Information Processing Systems}, pp.\
  13--23, 2019.

\bibitem[Lu et~al.(2020)Lu, Goswami, Rohrbach, Parikh, and Lee]{Lu2020}
Jiasen Lu, Vedanuj Goswami, Marcus Rohrbach, Devi Parikh, and Stefan Lee.
\newblock 12-in-1: Multi-task vision and language representation learning.
\newblock In \emph{Proceedings of the IEEE/CVF Conference on Computer Vision
  and Pattern Recognition}, pp.\  10437--10446, 2020.

\bibitem[Miech et~al.(2019)Miech, Zhukov, Alayrac, Tapaswi, Laptev, and
  Sivic]{Miech2019}
Antoine Miech, Dimitri Zhukov, Jean-Baptiste Alayrac, Makarand Tapaswi, Ivan
  Laptev, and Josef Sivic.
\newblock Howto100m: Learning a text-video embedding by watching hundred
  million narrated video clips.
\newblock In \emph{Proceedings of the IEEE international conference on computer
  vision}, pp.\  2630--2640, 2019.

\bibitem[Papineni et~al.(2002)Papineni, Roukos, Ward, and Zhu]{Papineni2002}
Kishore Papineni, Salim Roukos, Todd Ward, and Wei-Jing Zhu.
\newblock Bleu: a method for automatic evaluation of machine translation.
\newblock In \emph{Proceedings of the 40th annual meeting of the Association
  for Computational Linguistics}, pp.\  311--318, 2002.

\bibitem[Paszke et~al.(2019)Paszke, Gross, Massa, Lerer, Bradbury, Chanan,
  Killeen, Lin, Gimelshein, Antiga, et~al.]{Paszke2019}
Adam Paszke, Sam Gross, Francisco Massa, Adam Lerer, James Bradbury, Gregory
  Chanan, Trevor Killeen, Zeming Lin, Natalia Gimelshein, Luca Antiga, et~al.
\newblock Pytorch: An imperative style, high-performance deep learning library.
\newblock \emph{arXiv preprint arXiv:1912.01703}, 2019.

\bibitem[Pratap et~al.(2020)Pratap, Sriram, Tomasello, Hannun, Liptchinsky,
  Synnaeve, and Collobert]{Pratap2020}
Vineel Pratap, Anuroop Sriram, Paden Tomasello, Awni Hannun, Vitaliy
  Liptchinsky, Gabriel Synnaeve, and Ronan Collobert.
\newblock Massively multilingual asr: 50 languages, 1 model, 1 billion
  parameters.
\newblock \emph{arXiv preprint arXiv:2007.03001}, 2020.

\bibitem[Richley \& Walters(2020)Richley and Walters]{SCST2020}
David Richley and Harriet Walters.
\newblock Clinical guidelines by consensus recommendations for {ECG} reporting
  standards and guidance., 2020.
\newblock URL
  \url{https://scst.org.uk/wp-content/uploads/2020/06/CS4v1.1-ECG_Reporting_Guidelines_June-2020.pdf}.

\bibitem[Singh et~al.(2020)Singh, Goswami, and Parikh]{Singh2020}
Amanpreet Singh, Vedanuj Goswami, and Devi Parikh.
\newblock Are we pretraining it right? digging deeper into visio-linguistic
  pretraining.
\newblock \emph{arXiv preprint arXiv:2004.08744}, 2020.

\bibitem[Strodthoff et~al.(2020)Strodthoff, Wagner, Schaeffter, and
  Samek]{Strodthoff2020}
Nils Strodthoff, Patrick Wagner, Tobias Schaeffter, and Wojciech Samek.
\newblock Deep learning for {ECG} analysis: Benchmarks and insights from
  {PTB-XL}.
\newblock \emph{arXiv preprint arXiv:2004.13701}, 2020.

\bibitem[Sun et~al.(2019)Sun, Myers, Vondrick, Murphy, and Schmid]{Sun2019}
Chen Sun, Austin Myers, Carl Vondrick, Kevin Murphy, and Cordelia Schmid.
\newblock Videobert: A joint model for video and language representation
  learning.
\newblock In \emph{Proceedings of the IEEE International Conference on Computer
  Vision}, pp.\  7464--7473, 2019.

\bibitem[Tevet \& Berant(2020)Tevet and Berant]{Tevet2020}
Guy Tevet and Jonathan Berant.
\newblock Evaluating the evaluation of diversity in natural language
  generation.
\newblock \emph{arXiv preprint arXiv:2004.02990}, 2020.

\bibitem[Wagner et~al.(2020)Wagner, Strodthoff, Bousseljot, Samek, and
  Schaeffter]{Wagner2020}
Patrick Wagner, Nils Strodthoff, Ralf-Dieter Bousseljot, Wojciech Samek, and
  Tobias Schaeffter.
\newblock {PTB-XL}, a large publicly available electrocardiography dataset,
  2020.
\newblock URL \url{https://physionet.org/content/ptb-xl/1.0.1/}.

\bibitem[Wang et~al.(2018)Wang, Peng, Lu, Lu, and Summers]{Wang2018}
Xiaosong Wang, Yifan Peng, Le~Lu, Zhiyong Lu, and Ronald~M Summers.
\newblock Tienet: Text-image embedding network for common thorax disease
  classification and reporting in chest x-rays.
\newblock In \emph{Proceedings of the IEEE conference on computer vision and
  pattern recognition}, pp.\  9049--9058, 2018.

\bibitem[Willems et~al.(1991)Willems, Abreu-Lima, Arnaud, van Bemmel, Brohet,
  Degani, Denis, Gehring, Graham, van Herpen, et~al.]{Willems1991}
Jos~L Willems, Cassiano Abreu-Lima, Pierre Arnaud, Jan~H van Bemmel, Christian
  Brohet, Rosanna Degani, Bernard Denis, J{\"u}rgen Gehring, Ian Graham, Gerard
  van Herpen, et~al.
\newblock The diagnostic performance of computer programs for the
  interpretation of electrocardiograms.
\newblock \emph{New England Journal of Medicine}, 325\penalty0 (25):\penalty0
  1767--1773, 1991.

\bibitem[Xu et~al.(2015)Xu, Ba, Kiros, Cho, Courville, Salakhudinov, Zemel, and
  Bengio]{Xu2015}
Kelvin Xu, Jimmy Ba, Ryan Kiros, Kyunghyun Cho, Aaron Courville, Ruslan
  Salakhudinov, Rich Zemel, and Yoshua Bengio.
\newblock Show, attend and tell: Neural image caption generation with visual
  attention.
\newblock In \emph{International conference on machine learning}, pp.\
  2048--2057, 2015.

\bibitem[Yoon et~al.(2019)Yoon, Lee, Kim, Jeong, and Kang]{Yoon2019}
Wonjin Yoon, Jinhyuk Lee, Donghyeon Kim, Minbyul Jeong, and Jaewoo Kang.
\newblock Pre-trained language model for biomedical question answering.
\newblock In \emph{Joint European Conference on Machine Learning and Knowledge
  Discovery in Databases}, pp.\  727--740. Springer, 2019.

\bibitem[Zeng et~al.(2020)Zeng, Wen, Liu, and Qi]{Zeng2020}
Xianhua Zeng, Li~Wen, Banggui Liu, and Xiaojun Qi.
\newblock Deep learning for ultrasound image caption generation based on object
  detection.
\newblock \emph{Neurocomputing}, 392:\penalty0 132--141, 2020.

\bibitem[Zhang et~al.(2020{\natexlab{a}})Zhang, Williams, Titov, and
  Sennrich]{Zhang2020MNMT}
Biao Zhang, Philip Williams, Ivan Titov, and Rico Sennrich.
\newblock Improving massively multilingual neural machine translation and
  zero-shot translation.
\newblock \emph{arXiv preprint arXiv:2004.11867}, 2020{\natexlab{a}}.

\bibitem[Zhang et~al.(2020{\natexlab{b}})Zhang, Jiang, Miura, Manning, and
  Langlotz]{Zhang2020}
Yuhao Zhang, Hang Jiang, Yasuhide Miura, Christopher~D Manning, and Curtis~P
  Langlotz.
\newblock Contrastive learning of medical visual representations from paired
  images and text.
\newblock \emph{arXiv preprint arXiv:2010.00747}, 2020{\natexlab{b}}.

\bibitem[Zhu et~al.(2018)Zhu, Lu, Zheng, Guo, Zhang, Wang, and Yu]{Zhu2018}
Yaoming Zhu, Sidi Lu, Lei Zheng, Jiaxian Guo, Weinan Zhang, Jun Wang, and Yong
  Yu.
\newblock Texygen: A benchmarking platform for text generation models.
\newblock In \emph{The 41st International ACM SIGIR Conference on Research \&
  Development in Information Retrieval}, pp.\  1097--1100, 2018.

\end{thebibliography}
\bibliographystyle{iclr2021_conference}

\appendix

\clearpage

\begin{subappendices}
\renewcommand{\thesubsection}{\Alph{section}.\arabic{subsection}}
\section{Datasets}
\label{appendix:datasets}

\subsection{Data Preprocessing}
\label{appendix:data_description}

The ECG frames consisted of 2500 samples and consecutive frames had no overlap with one another. Data splits were always performed at the patient-level.

\textbf{PTB-XL} \citep{Wagner2020}. Each ECG recording was originally 10 seconds with a sampling rate of 500Hz. We extract 5-second non-overlapping segments of each recording generating frames of length 2500 samples. We follow the diagnostic class labelling setup suggested by \cite{Strodthoff2020} which resulted in five classes: Conduction Disturbance (CD), Hypertrophy (HYP), Myocardial Infarction (MI), Normal (NORM), and Ischemic ST-T Changes (STTC). Furthermore, we only consider ECG segments with one label assigned to them. The ECG frames were standardized to follow a standard Gaussian distribution. 

\subsection{Data Samples}
\label{appendix:instances}

In this section, we outline the number of instances used during training.

\begin{table}[h]
\centering
\caption{Number of instances (number of patients) used during training. These represent sample sizes for all 12 leads.}
\label{table:data_splits}
\begin{tabular}{c | c c c}
\toprule
Dataset & Train & Validation & Test\\
\midrule
\multirow{1}{*}{PTB-XL}& 22,670 (11,335) &3,284 (1,642)&3,304 (1,152)\\
\bottomrule
\end{tabular}
\end{table}

\subsection{Vocabulary Tokens}
\label{appendix:tokens}

In this section, we outline the number of language-specific tokens available in each language's vocabulary for the two datasets. 

\begin{table}[h]
\centering
\caption{Number of language-specific tokens in each dataset}
\label{table:tokens}
\begin{tabular}{c | c c c c c c c c c }
\toprule
Dataset & de & el & en & es & fr & it & pt \\
\midrule
\multirow{1}{*}{PTB-XL}& 2206 & 2662 & 1606 & 1950 & 1974 & 1866 & 2010 \\
\bottomrule
\end{tabular}
\end{table}

\clearpage

\section{Translation Details}
\label{appendix:physiological_captioning_translation}

In this section, we outline the steps taken to translate the ECG reports originally found in the PTB-XL dataset. We remind readers that although these ECG reports are a mixture of English and German, they are predominantly in the latter. As a result, we treat German as the source language from which we translate the reports to other languages. More specifically, we follow these steps.  

\begin{enumerate}
    \item  We leverage the Google Translate API to first detect the source language of each ECG report. Although the majority of the reports are in German, some are in English, and this language detection step ensures that the ultimate translation is of a higher quality. 
    \item We continue to leverage the Google Translate API to translate ECG reports from the identified source language to the target language of interest. 
    \item Due to imperfections in the Google Translate API, certain ECG reports may not be translated in full or translated at all. To minimize the incidence of such cases, we repeat Step 2 several times and stop once we reach the following criterion: we deploy the language detection module of the Google Translate API on the translated reports to confirm that over 90\% of them are indeed in the translated language. Although this implies that the final translated reports may have some noise, we found that this did not prevent our algorithm from learning appropriately. 
\end{enumerate}

\clearpage

\section{Implementation Details}
\label{appendix:physiological_captioning_implementation}

\subsection{Network Architectures}
\label{appendix:physiological_captioning_network}

In this section, we outline the neural network architectures used for our encoder and decoder. More specifically, we use the architecture shown in Table~\ref{table:encoder_architecture} for the encoder and that shown in Table~\ref{table:decoder_architecture} for the decoder. 

\begin{table}[!h]
\footnotesize
\centering
\caption{Encoder architecture used for experiments conducted on the PTB-XL dataset. \textit{K}, \textit{C}\textsubscript{in}, and \textit{C}\textsubscript{out} represent the kernel size, number of input channels, and number of output channels, respectively. A stride of 3 was used for all convolutional layers. $M$ represents the dimension of the final representation. We only use layer 5 when performing supervised pre-training. When captioning, layer 4 outputs $L$ temporal features.}
\label{table:encoder_architecture}
\begin{tabular}{c c c}
\toprule
Layer Number &Layer Components&Kernel Dimension\\
\midrule
	\multirow{5}{*}{1}&Conv 1D & 7 $\times$ 12 $\times$ 32 (\textit{K} $\times$ \textit{C}\textsubscript{in} $\times$ \textit{C}\textsubscript{out})\\
										& BatchNorm &\\
										& ReLU& \\
										& MaxPool(2)& \\
										& Dropout(0.1) &\\
	\midrule
	\multirow{5}{*}{2}&Conv 1D & 7 $\times$ 32 $\times$ 64\\
										& BatchNorm& \\
										& ReLU &\\
										& MaxPool(2) &\\
										& Dropout(0.1)& \\
	\midrule
	\multirow{5}{*}{3}&Conv 1D & 7 $\times$ 64 $\times$ 128 \\
										& BatchNorm &\\
										& ReLU &\\
										& MaxPool(2) &\\
										& Dropout(0.1) &\\
	\midrule
	\multirow{2}{*}{4}&Linear&128 $\times$ $M$ \\
										& ReLU &\\
	\midrule
	\multirow{1}{*}{5}&Linear &$M$ $\times$ C (classes) \\
\bottomrule
\end{tabular}
\end{table}

\begin{table}[!h]
\footnotesize
\centering
\caption{Decoder architecture used for experiments conducted on the PTB-XL dataset. $E=300$ represents the dimension of the representations from the encoder and the representations of the decoder tokens. $H=4$ represents the number of heads used in each of the self and cross-attention modules. C\textsubscript{lang} represents the number of tokens in a specific language.}
\label{table:decoder_architecture}
\begin{tabular}{c c c}
\toprule
Layer Number &Layer Components&Kernel Dimension\\
\midrule
	\multirow{1}{*}{1}&Transformer Decoder Layer & $E$, $H$ \\
	\midrule
	\multirow{1}{*}{2}&Transformer Decoder Layer & $E$, $H$ \\
	\midrule
	\multirow{1}{*}{3}&Transformer Decoder Layer & $E$, $H$ \\
	\midrule
	\multirow{1}{*}{4}&Transformer Decoder Layer & $E$, $H$ \\
	\midrule
	\multirow{1}{*}{5}&Linear &$E$ x C\textsubscript{lang} \\
\bottomrule
\end{tabular}
\end{table}

\begin{table}[!h]
\small
\centering
\caption{Batchsize and learning rates used for training. The Adam optimizer was used for all experiments.}
\label{table:learning_rates}
\begin{tabular}{c | c c }
\toprule
Stage & Batchsize & Learning Rate\\
\midrule
\multicolumn{3}{c}{\textit{Encoder}} \\
\midrule
Supervised Pre-training & 128 & 10\textsuperscript{-5} \\
\midrule
\multicolumn{3}{c}{\textit{Decoder}} \\
\midrule
MLM Pre-training & 128 & 10\textsuperscript{-3} \\
ELECTRA Pre-training & 128 & 10\textsuperscript{-3} \\
RTLP Pre-training & 128 & 10\textsuperscript{-3} \\
MARGE Pre-training & 64 & 10\textsuperscript{-4} \\
\midrule
\multicolumn{3}{c}{\textit{Combined}} \\
\midrule
Fine-tuning & 128 & 10\textsuperscript{-3} \\
\bottomrule
\end{tabular}
\end{table}

\newpage

\subsection{Encoder Pre-training}
\label{appendix:physiological_captioning_encoder_pretraining}

In this section, we outline the task used to pre-train the encoder of the captioning system in a supervised manner. Specifically, we learn an encoder, $f_{\theta}: u \in \mathbb{R}^{P \times D} \rightarrow y \in \mathbb{R}^{C}$ parameterized by $\theta$, that maps $P=12$ $D$-dimensional ECG signals, $u$, (where $P$ represents the number of leads) to a $C$-dimensional output representing the probability assigned to each of the cardiac arrhythmia classes. When leveraging the PTB-XL dataset, $C=5$. For a mini-batch of size, $B$, and where $c_{i}$ represents the ground-truth class for a particular instance, $x_{i}$, we learn this behaviour by optimizing the following categorical cross-entropy loss. 
\begin{equation}
    \mathcal{L_{CE}} = - \frac{1}{B} \sum_{i=1}^{B} \log p_{\theta}(y_{i} = c_{i}) 
\end{equation}

We checkpoint, and eventually exploit, the parameters, $\theta$, that coincide with the lowest loss observed on the validation set. This ensures that we use parameters that do not exhibit overfitting. 

\newpage

\subsection{Baseline Implementations}
\label{appendix:physiological_captioning_baseline_implementations}

\textbf{Masked Language Modelling.} Masked language modelling (MLM) can be thought of as analogous to a denoising autoencoder. Inputs are perturbed and the network is tasked with generating the original, unperturbed version of the input. In the context of natural language processing, a fraction $F=0.15$ of the tokens in a sentence are chosen to be masked. Of these chosen tokens, $80\%$ are replaced with the token $\mathrm{[MASK]}$, $10\%$ are replaced with a random token from the vocabulary, and the final $10\%$ are not replaced at all. The motivation behind this task lies in the ability of the network to leverage the context of masked tokens to correctly predict them. This, in turn, allows for the learning of rich representations. In our context, and to allow for a fair comparison to the multilingual pre-training methods, we follow the original implementation introduced by \cite{Devlin2018} for each of the language mini-batches. More specifically, at each iteration, we load $N$ mini-batches corresponding to $N$ languages and perform MLM on each of these batches. 

\textbf{ELECTRA.} ELECTRA, as opposed to MLM introduced above, is a discriminative language representation learning method. ELECTRA builds upon the implementation of MLM in the following ways. First, instead of masking tokens and tasking the network with generating the original token, ELECTRA performs a binary classification of whether a token was replaced or not. The motivation for doing so lies in the alleged unnecessary complexity associated with generative language representation learning methods. Moreover, instead of replacing tokens with the $\mathrm{[MASK]}$ token, ELECTRA proposes to do so by exploiting the predicted outputs of an MLM. This increases the likelihood that replaced tokens are in-distribution. As a result, ELECTRA simultaneously trains an MLM network and a binary classifier. In our context, we follow the original implementation introduced by \cite{Clark2020} for each of the $N$ mini-batches. 

\textbf{MARGE.} MARGE is a generative multilingual language representation learning method that exploits source documents in various languages to generate text from a similar yet distinct target document. For example, $M$ source documents with $M*S$ tokens are encoded and leveraged by a decoder to generate the $T$ tokens in the target document. In doing so, the network is able to capture relationships between languages and thus learn representations useful for downstream multilingual tasks. In the original implementation \citep{Lewis2020}, similar documents need to be retrieved from a database. In our context, however, our ECG reports are available in $N$ different languages and thus the target document is formed by a report in one language and the source documents are formed by reports in the remaining $N-1$ languages. Since our ECG reports were translated from a single original language, we used reports in this language as target documents. For PTB-XL, this amounts to using German.

\end{subappendices}

\end{document}